\documentclass[journal]{IEEEtran}
\usepackage{amsmath,amsfonts}
\usepackage{epsfig}
\usepackage{epstopdf}
\usepackage{algorithmic}
\usepackage[ruled,linesnumbered]{algorithm2e}
\usepackage{array}
\usepackage{subfigure}
\usepackage{textcomp}
\usepackage{stfloats}
\usepackage[T1]{fontenc}
\usepackage{aecompl}
\usepackage{booktabs}
\usepackage{siunitx}
\usepackage{amssymb}
\usepackage{multirow}
\usepackage{diagbox}
\usepackage{balance}
\usepackage{url}
\usepackage{verbatim}
\usepackage{graphicx}

\def\BibTeX{{\rm B\kern-.05em{\sc i\kern-.025em b}\kern-.08em
    T\kern-.1667em\lower.7ex\hbox{E}\kern-.125emX}}
\usepackage{balance}
\begin{document}
\title{\LARGE \bf
CEASE: Collision-Evaluation-based Active Sense System for Collaborative Robotic Arms}
\author{Xian Huang, Yuanjiong Ying and Wei Dong
\thanks{Manuscript received: AA; Revised: AA; Accepted AA. This paper was recommended for publication by Editor AA upon evaluation of the Associate Editor and Reviewers’ comments. This work was supported in part by the National Natural Science Foundation of China Grant 51975348 and in part by Shanghai Rising-Star Program under Grant 22QA1404400. \emph{(Corresponding author: Wei Dong.)}}

\thanks{All authors are with the State Key Laboratory of Mechanical System and Vibration, School of Mechanical Engineering, Shanghai Jiao Tong University,  Shanghai 200240, China (e-mail: hx2020@sjtu.edu.cn; dr.dongwei@sjtu.edu.cn).}
\thanks{Digital Object Identifier (DOI): see top of this page.}}

\markboth{Journal of \LaTeX\ Class Files,~Vol.~18, No.~9, September~2020}%
{CEASE: Collision Estimation Active Sense System for Collaborative Robotic Arms}
\maketitle
\thispagestyle{empty}
\pagestyle{empty}

\begin{abstract}
 Collision detection via visual fences can significantly enhance the safety of collaborative robotic arms. Existing work typically performs such detection based on pre-deployed stationary cameras outside the robotic arm's workspace. These stationary cameras can only provide a restricted detection range and constrain the mobility of the robotic system. To cope with this issue, we propose an active sense method enabling a wide range of collision risk evaluation in dynamic scenarios. First, an active vision mechanism is implemented by equipping cameras with additional degrees of rotation. Considering the uncertainty in the active sense, we design a state confidence envelope to uniformly characterize both known and potential dynamic obstacles. Subsequently, using the observation-based uncertainty evolution, collision risk is evaluated by the prediction of obstacle envelopes. On this basis, a Markov decision process was employed to search for an optimal observation sequence of the active sense system, which enlarges the field of observation and reduces uncertainties in the state estimation of surrounding obstacles.  Simulation and real-world experiments consistently demonstrate a 168\% increase in the observation time coverage of typical dynamic humanoid obstacles compared to the method using stationary cameras, which underscores our system’s effectiveness in collision risk tracking and enhancing the safety of robotic arms.
\end{abstract}


\section{INTRODUCTION}

\IEEEPARstart{R}{obotic} arms play a vital role in modern society, as the widespread use of automation technology continues to grow. To enhance the flexibility and maneuverability of robotic arms, an increasing number of factories are adopting collaborative robotic arms \cite{CR2021}. Collaborative robots inherently require sustained human interaction, which highlights the necessity for robust obstacle avoidance capabilities in the context of robotic arms \cite{CRCA}. To ensure robust obstacle avoidance, sufficient environmental sensing is necessary \cite{CRV}. 
\par Robots can perceive their surroundings through various methods \cite{skin,RGB} and RGB-D vision is particularly favored due to its lightweight and flexibility \cite{RGBD,POCT}. Nevertheless, the effectiveness of this perception method is often limited by the camera's field of view (FOV). In vision-based robotic arm collision avoidance, the camera is commonly stationary \cite{avoid1,RGBDC}, causing the robotic arm to focus solely on obstacles within its frontal view while neglecting obstacles on the sides. This limitation increases the risk of collisions with dynamic obstacles. Although robotic arms can plan within known spaces \cite{avoid1}, the solution space is significantly reduced, possibly rendering it unable to find feasible paths.
\par Drawing inspiration from human behavior, individuals frequently survey their surroundings while navigating intersections or crosswalks \cite{peds1}. Compared with aforementioned robotic vision technique, this behavior is a proactive measure to anticipate the sudden appearance of obstacles from the sides \cite{peds3}. Additionally, when we identify the presence of a dynamic obstacle, such as a car, we maintain vigilant observation to preemptively avoid potential accelerations or swerving that might lead to a collision \cite{peds2}. The overarching objective is to maximize certainty regarding the state of obstacles, thereby developing avoidance strategies in time. 
\par In this letter, we designed a \textbf{C}ollision-\textbf{E}valuation-based \textbf{A}ctive \textbf{SE}nse system called CEASE system with active vision mechanisms, facilitating additional degrees of rotation for the RGB-D camera through the utilization of two mutually perpendicular servos. We first propose a unified state confidence envelope (SCE) to quantify the level of certainty associated with the states of dynamic obstacles. On this basis, we formulated an observation-based uncertainty evolution (OUE) law to depict the evolution of both known and potential SCEs under a specific observation over time. Subsequently, the Collision-free Optimal Observation Sequence (COOS) search method using Markov decision process is raised to identify the direction of the maximum collision risk of the robotic arm over time. Eventually, dynamic obstacles tracking is facilitated and collision risk is reduced by the continuous scheduling of the active vision mechanisms, which enhances the safety of collaborative robotic arms. We conduct thorough tests in both simulation and real-world scenarios to validate our proposed method.
\par The main contributions of this letter are as follows:
\begin{enumerate}
    \item An SCE is proposed to uniformly quantify states of general dynamic obstacles. 
    \item An OUE and associated data structure designed to signify the potential risk.
    \item A COOS search approach for an active vision mechanism to ensure safety in robotic arm execution, which enhance collision probability estimation efficiently.
\end{enumerate}

\section{RELATED WORK}
In the operation of a collaborative robotic arm, it is crucial to guarantee the absence of collisions between the operators and the robotic arm. Because of the dynamic movement of the operators, solely relying on offline path planning poses challenges for the collaborative robotic arm to effectively accomplish collision-free tasks \cite{avoid2}. Consequently, collaborative robotic arms commonly employ sensors to observe the environment continuously, ensuring the safety. For example, Ge \textit{et al.} \cite{colla1} employs an electronic skin structure to envelop the robotic arm, enabling it to detect distances between obstacles and itself. Furthermore, Safeea \textit{et al.} \cite{avoid2} employs a multi-sensor fusion strategy for dynamic obstacle detection to generate obstacle avoidance paths according to offline path.
\par Collaborative robotic arms often employ RGB-D cameras for collision detection. The cameras are typically mounted in either the ``eye in hand" or ``eye out of hand" configurations. The former, where the camera base is mobile, allows for extensive exploration of a broader area. This ``eye in hand" style has been successfully employed in diverse tasks such as plant picking \cite{vision1} and real-time obstacle avoidance \cite{vision2}. However, this approach, integrating camera planning with robotic arm planning, often makes it hard to strike a balance between task execution efficiency and observation quality. On the other hand, the latter ``eye out of hand" style involves mounting a fixed camera with a static viewing area, typically employed for smaller-scale tasks like object grasping recognition \cite{vision3,vision4} and robotic arm calibration \cite{vision5}. In such collaborative tasks, an external camera is frequently utilized to monitor the workspace and prevent collisions between the robotic arm and humans \cite{avoid1,avoid3}. Nevertheless, this mounting style requires the camera to be positioned higher than the robotic arm to ensure unobstructed vision within FOV during arm movements, which leads to low mobility and high price.
\par Due to FOV constraints, the quality of RGB-D camera observations varies depending on the observation direction and camera base position. In exploration tasks, given the constraints of the FOV, a robot equipped with an RGB-D camera requires careful path planning to find the best position observing the environment. To address this problem, the Next Best View (NBV) algorithm \cite{NBV1} was proposed. This algorithm defines the robot's trajectory points in conjunction with camera directions as viewpoints, incorporating a corresponding information gain value \cite{NBV2}, which is called the viewpoint value. Zhou \textit{et al}. \cite{NBV3} further refined the algorithm by integrating a graph search algorithm, enhancing the exploration's optimality. The concept of viewpoints extends to 3D reconstruction tasks and irregular object reconstruction tasks as well \cite{NBV4}. However, the majority of the aforementioned studies focus on exploration tasks, leaving the construction method of information gain for obstacle avoidance tasks yet to be thoroughly investigated.
\par The implementation of active vision can significantly reduce the challenges associated with path planning and improve the quality of observations. Pan et al. \cite{drone5} devised a rotation mechanism for the camera to track people on the move, facilitating the decoupling of the UAV yaw angle and the camera angle. Moreover, In obstacle avoidance tasks, the dynamics of yaw angle rotation limits the observation direction in UAV. In response, Chen et al. \cite{drone4} proposed a rotation mechanism and an observation planning strategy to optimize the camera angle based on the viewpoint value at each angle. This innovative approach effectively overcomes the FOV limitation and provides a robust solution for dynamic obstacle avoidance. However, the lack of long-term planning can lead to optimal local solutions.
\par In uncertain environments, the POMDP \cite{MDP1,MDP2} is often used to solve problems that combine observation with path planning. The POMDP framework facilitates the evolution of the robot's internal state through observation, employing rewards to optimize both perception and movement \cite{MDP3}. The set of states within the Markov process is defined as the set of viewpoints, and the optimization of rewards facilitates active vision planning\cite{MDP2}. However, previous work has not addressed the real-time performance of highly nonlinear obstacle avoidance tasks in uncertain environments. A real-time scheduling method remains an essential aspect yet to be explored.

\section{PROBLEM FORMULATION}

\begin{figure}[!t]
    \includegraphics[width=9cm]{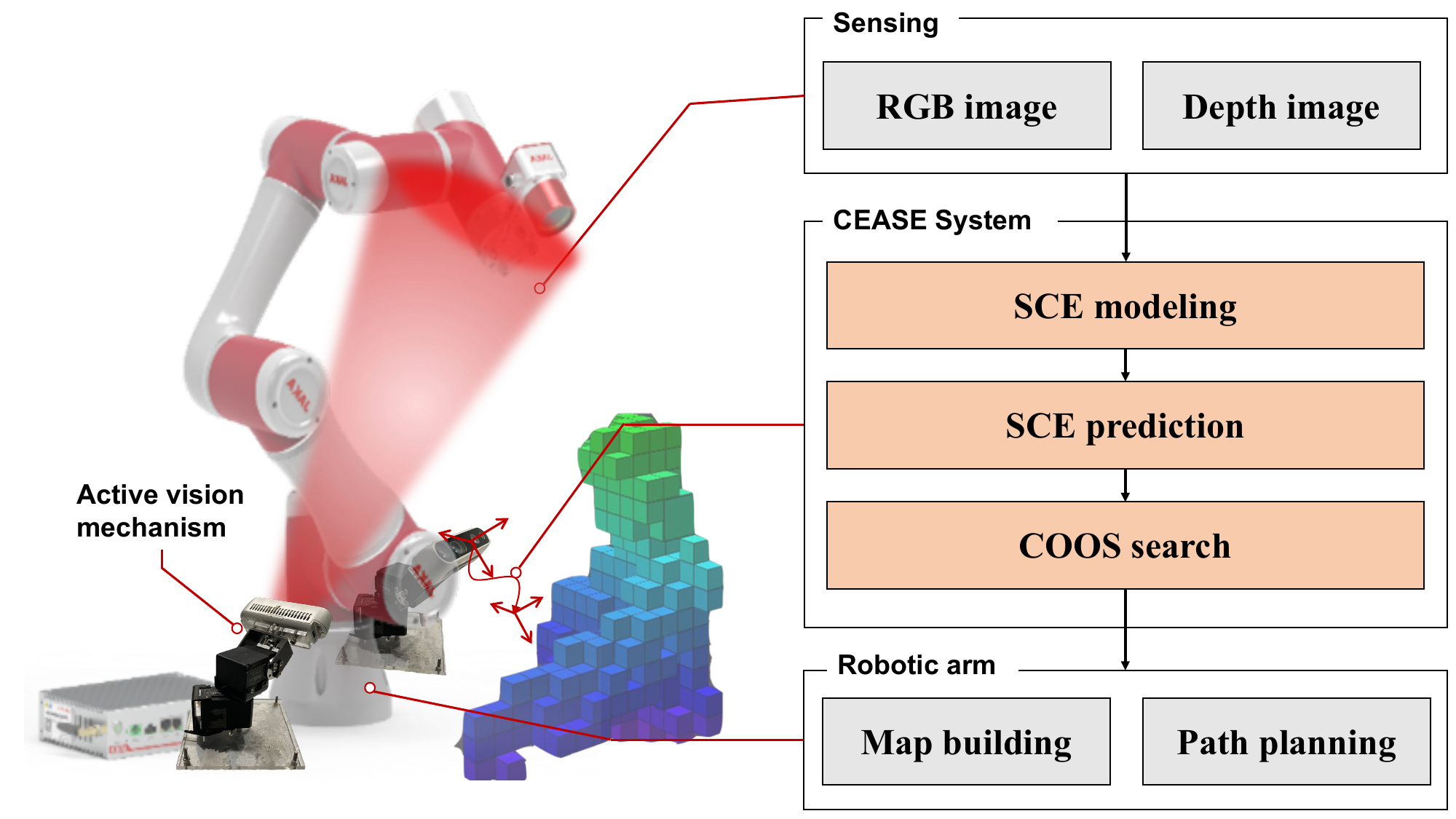}
    \centering
    \caption{Overview of our CEASE system.}
    \label{fig1}
\end{figure}

Our active vision mechanism requires the capability to observe the entire workspace. To fulfill this requirement, we designed an active vision mechanism comprising two mutually perpendicular servos and an RGB-D camera, as illustrated in Fig. \ref{fig1}. Within the CEASE system, we position two of these active vision mechanisms on both sides of the robotic arm to ensure that the cameras' FOV remains unobstructed by the robotic arm, thereby enhancing the detection of dynamic obstacles.

\par We aim to devise an algorithm that allows our active vision mechanism to determine the optimal direction for obstacle observation over a given time period, thereby reducing the collision risk between the obstacles and the robotic arm. The problem can be formulated as follows: Given the initial vision state \(\boldsymbol{s}_{v}^0 = [\theta_1, \theta_2] \in \mathbb{S}^2\) of the active vision mechanism, along with the subsequent robot trajectory \(\tau:[0,t] \mapsto \boldsymbol{q}_{free}\), the trajectory of the active vision mechanism \(\tau_v:[0,t] \mapsto \mathbb{S}^2\) is designed to minimize the collision probability estimation (CPE) throughout this time interval

\begin{equation}
    \tau_{v,d} = \arg_{\tau_v} \max \left(1-\hat{p}^{0 \sim  t}_c \right)
\end{equation}
where \(\hat{p} ^{0 \sim  t}_c\) represents the CPE between robotic arm and obstacles and $\theta_1, \theta_2$ are the angles of the mutually perpendicular servos. Given the challenge of obtaining the collision probability density function, it becomes necessary to discretize the problem. Consequently, we divide the time interval \([0, t]\) into \(n\) segments \([0, t_1],\dots , [t_{i-1},t_i], \dots, [t_{n-1},t_n]\), transforming our problem into the following formulation
\begin{equation}
    \tau_{v,d} = \arg_{\tau_v} \max{\sum_{i=1}^n{\gamma^i \ln(1-\hat{p} ^{i,i+1}_{c})}}\label{eq:optimal}
\end{equation}
where \(\gamma\) is a decay coefficient less than 1, \({p} ^{i,i+1}_{c}\) represents the probability of collision between the robotic arm and the obstacles in the time interval \([t_{i-1},t_i]\), and \(\hat{p} ^{i,i+1}_{c}\) is its corresponding estimation. Subsequently, the trajectory of the active vision mechanism that we obtain is discretized into \(n\) segments. In other words, the state of the active vision mechanism in the \(i\)-th time interval \([t_{i-1}, t_i]\) should be \(\boldsymbol{s}_{v}^{i-1} \in \mathbb{S}^2\).

\section{COLLISION-EVALUATION-BASED ACTIVE SENSE SYSTEM}
In this section, we begin by providing an overview of our system, and proceed to a unified model SCE for characterizing obstacle certainty and state. Subsequently, we propose the OUE law to depict the evolution of collision risk according to observed SCEs and potential obstacles. Eventually, We address the optimization of the observation sequence based on the aforementioned OUE law.
\subsection{System Design}
Utilizing the framework established for the active vision mechanism and the associated problem, the overall structure of our system is illustrated in Fig. \ref{fig1}. Following the environmental sensing by the camera on the active vision mechanism, our CEASE system processes RGB and depth information to construct the SCE. Subsequently, discrete-time obstacle prediction is executed based on this model, leading to the formulation of an optimization problem using (\ref{eq:optimal}). This optimization aims to identify the COOS and execute camera angle control. Additionally, octree map for all obstacles were generated, facilitating robotic arm obstacle avoidance based on the map.
\subsection{SCE Modeling}
There exist various types of dynamic obstacles in the operational environment of robotic arms. However, humanoid obstacles are critical in the collaborative robotic arm scenario, where interference between humans and robotic arms is common. Hence, our objective is to construct a model that represents humanoid dynamic obstacles. Consequently, we employ a methodology capable of estimating the pose and velocity of humanoid obstacles within the depth camera's FOV. Due to inherent errors in the sensor's observation estimates, we use a covariance matrix to characterise such an estimation error.
\par We conceptualize the humanoid dynamic obstacle as an SCE, fully characterized by the obstacle state $\boldsymbol{s}_e = [\boldsymbol{x_e}^T, \boldsymbol{o_e}^T,\boldsymbol{v_e}^T,\boldsymbol{\omega_e}^T,u_e] \in \mathbb{R}^{13}$ and obstacle covariance matrix $\Sigma_{e} = [\Sigma_x^T, \Sigma_o^T, \Sigma_v^T, \Sigma_{\omega}^T]^T  \in \mathbb{R}^{12 \times 3}$. $\boldsymbol{s}_e$ consists of obstacle position $\boldsymbol{x_e}^T $, obstacle rotation vector $\boldsymbol{o_e}^T $, obstacle velocity $\boldsymbol{v_e}^T $, obstacle angular velocity $\boldsymbol{\omega_e}^T $ and obstacle certainty $ u $. Specifically, we define the concept of certainty as the probability that the state of the observed object $\boldsymbol{s}_{er}$ is in the neighborhood of the estimated state $\hat{\boldsymbol{s}}_{e}$ under a single observation. For a state $\boldsymbol{s}_{e0}$, if $\| \boldsymbol{x}_{e0}^T - \boldsymbol{x}_{e}^T \| < r_p$ and $\| \boldsymbol{o}_{e0}^T - \boldsymbol{o}_{e}^T \| < r_o$, then we define $\boldsymbol{s}_{e0}$ to be in the neighborhood of $\boldsymbol{s}_{e}$. 
\par By definition, certainty is expressed as a conditional probability, denoted as $u = p\left(\overline{\boldsymbol{s}_e}  \mid \mathcal{O}, \boldsymbol{s}_{el} \right)$. Here, $\overline{\boldsymbol{s}_e}$ does not denote a specific state but an event, specifically, that this state $\boldsymbol{s}_e$ is in the neighborhood of the real state $\boldsymbol{s}_{er}$ of the dynamic obstacle, and $\boldsymbol{s}_{el}$ represents the state of this obstacle at the last discrete moment in time. Since all subsequent discussions are conditional on the previous state being $\boldsymbol{s}_{el}$, we can abbreviate $u = p\left(\overline{\boldsymbol{s}_e} \mid \mathcal{O}, \boldsymbol{s}_{el} \right)$ as $u = p\left(\overline{\boldsymbol{s}_e }\mid \mathcal{O} \right)$. Observation $\mathcal{O}: \boldsymbol{s}_e \mapsto \{\text{0, 1}\}$ is a mapping, which can be categorized into valid and invalid observations. Valid observation implies that the vision state facilitates the estimation of the obstacle's state, where $\mathcal{O} = \text{1}$. Conversely, an invalid observation occurs when the state of the obstacle cannot be estimated due to occlusion or being beyond the FOV, where $\mathcal{O} = \text{0}$.

\subsection{OUE law}
Our OUE law is designed to provide a comprehensive prediction of both known and potential SCEs within the environment under a specific observation. We denote the set of all known SCEs at \textit{i}-th time as
$\mathcal{E}^i$ then the OUE law can be expressed as
\begin{equation}
    \mathcal{E}^{i+1}=\mathbb{T}(\mathcal{E}^{i}, \mathbb{SR}^i, \mathcal{O} )
\end{equation}
where $\mathbb{T}(\mathcal{E}^{i}, \mathbb{SR}^{i+1}, \mathcal{O})$ represent our OUE law and $\mathbb{SR}^i$ is a data structure representing the evolution law of potential obstacles which we will introduce later. 

\subsubsection{OUE law for known SCEs}
\par For known SCEs in $\mathcal{E}^{i}$, their position, orientation, velocity, and angular velocity evolve according to the following formula (\ref{eq:state_update})
\begin{equation}
\begin{aligned}
    \boldsymbol{x}_e^{i+1}(k) &= \boldsymbol{x}_e^{i}(k) + t\boldsymbol{v}_e^{i}(k)  \\
    \boldsymbol{o}_e^{i+1}(k) &= \cos{\theta}\boldsymbol{o}_e^{i}(k) + (1-\cos{\theta})(\boldsymbol{o}_e^{i}(k) \cdot \boldsymbol{r}^{i}(k))\boldsymbol{r}^{i}(k) \\
    &+ \sin{\theta}\cdot \boldsymbol{r}^{i}(k) \times \boldsymbol{o}_e^{i}(k) \\
    \boldsymbol{v}_e^{i+1}(k) &= \boldsymbol{v}_e^{i}(k) \\
    \boldsymbol{\omega}_e^{i+1}(k) &= \boldsymbol{\omega}_e^{i}(k)
\end{aligned}\label{eq:state_update}
\end{equation}
where $t = t^{i+1}-t^i$ represents the discrete time interval and $k$ represents the \textit{k}-th SCE in $\mathcal{E}^{i}$. The $\theta$ in (\ref{eq:state_update}) represents the angle of rotation, and $\theta = \left\lVert \boldsymbol\omega \right\rVert t$. $\boldsymbol{r}$ represents the unit vector in the $\boldsymbol\omega$ direction.

\par Based on the definition of $u$, we can formulate the update rule for certainty. If an observation is valid, then the probability that our estimated probability aligns with the actual state is as follows
\begin{equation}
    u=p(\overline{\boldsymbol{x}_e}^T) \cdot p(\overline{\boldsymbol{o}_e}^T)  
\end{equation}
where $\overline{\boldsymbol{x}_e}^T$ and $\overline{\boldsymbol{o}_e}^T$ represent the actual position $\boldsymbol{x}_{er}^T$ is in the neighborhood of $\boldsymbol{x}_e^T$, and the actual pose $\boldsymbol{o}_{er}^T$ is in the neighborhood of $\boldsymbol{o}_e^T$, respectively. Both position and pose follow three-dimensional Gaussian distributions, and the probabilities mentioned above can be computed based on the covariance matrix of the state estimation. However, in the case of an invalid observation, the evolution of the SCE will follow the natural evolution method shown in the equation below
\begin{equation}
    u=p(\overline{\boldsymbol{s}_e} )=p(\overline{\boldsymbol{s}_e} \mid \boldsymbol{s}_{el} )= p(\overline{\boldsymbol{x}_e^*}^T) \cdot p(\overline{\boldsymbol{o}_e^*}^T)
\end{equation}
\begin{equation}
    \boldsymbol{x}_e^* \sim \mathcal{N}\left(\boldsymbol{x}_{el}+\boldsymbol{v}_{el}t, \Sigma_x + t^2\Sigma_v \right)
\end{equation}
\begin{equation}
    \boldsymbol{o}_e^* \sim \mathcal{N}\left(\boldsymbol{o}_e, t^2\Sigma_\omega \right).
\end{equation}

\par Due to the high degree of nonlinearization of the transfer equation for the orientation, we consider only the Gaussian error in the angular velocity estimation for the calculation.
\par The position and velocity matrices in the covariance matrix of $k$-th SCE evolve according to the following (\ref{eq:covariance_update}). As orientation contributes minimally to collision detection and exhibits strong nonlinearity, their covariance matrix values are maintained as constants
\begin{equation}
\begin{aligned}
    \Sigma_v^{i+1}(k) & = \Sigma_v^i(k) + \left(1-\mathcal O(\boldsymbol{s}_{ek}) \right)t^2\boldsymbol{a}\boldsymbol{a}^T \\
    \Sigma_x^{i+1}(k) & = \Sigma_x^i(k)+\left(1-\mathcal O(\boldsymbol{s}_{ek}) \right)t^2 \Sigma_v^{i} (k)
\end{aligned}\label{eq:covariance_update}
\end{equation}
where $\boldsymbol{a} =a_{\text{max}} \cdot [1~ 1 ~ 1]^T$ and $a_{\text{max}}$ denote the priori maximum acceleration. This implies that invalid observations lead to the divergence of the prediction covariance matrix of the SCE.
\begin{figure}[!t]
    \includegraphics[width=9cm]{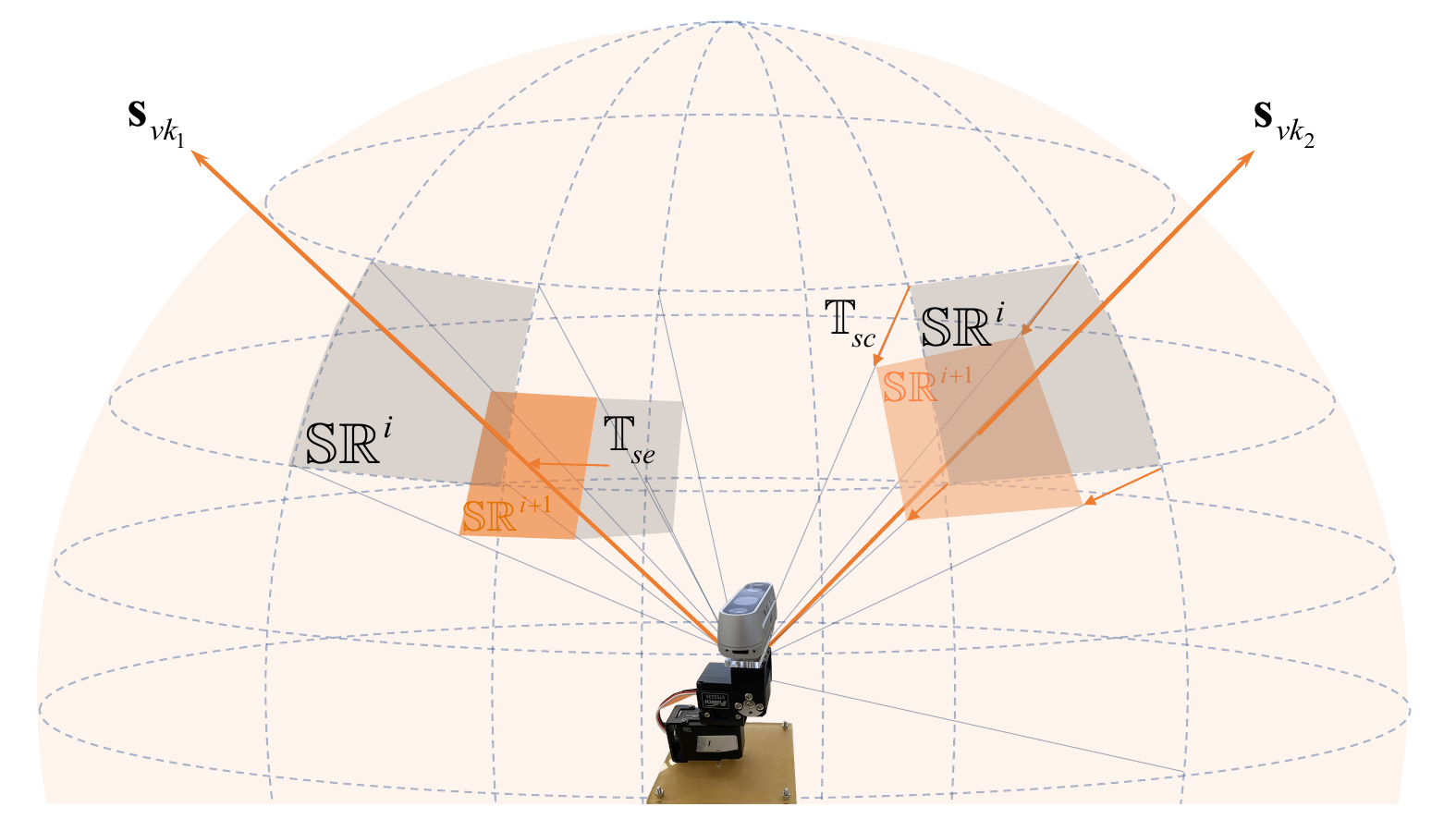}
    \centering
    \caption{Schematic representation delineating the safe region with the inclusion of the two evolutionary approaches, denoted as $\mathbb{T}_{se}$ and $\mathbb{T}_{sc}$.}
    \label{fig3}
\end{figure}
\par In conclusion, we depict the OUE law for known SCEs as follows
\begin{equation}
    \mathcal{E}^{i+1}_k= f(\mathcal{E}^{i},\mathcal{O}).
\end{equation}
\subsubsection{OUE law for potential SCEs}
\par In this subsection, we introduce the concept of a safety region, denoted as $\mathbb{SR}$, which is designed to characterize the movement of potential SCEs. The initial safe region $\mathbb{SR}^0$ is defined as a sphere with its center positioned at the termination of the first link of the active vision mechanism, and the sphere's radius corresponds to the depth of the camera's FOV
\begin{equation}
    \mathbb{SR}^0(\boldsymbol{s}_{vk}) = d_{max},\quad \boldsymbol{s}_{vk} \in\{\boldsymbol{s}_{v1}, \boldsymbol{s}_{v2}, \cdots, \boldsymbol{s}_{vn_{sf}}\}
\end{equation}
where  $n_{sf}$ represents the number of discrete vision state.
\par Employing spherical coordinates, we partition the entire space and employ an array $\mathbb{SR}^i(\boldsymbol{s}_{vk})$ to document the depth that remains within the current safety region at each discrete vision state, derived from the partition. The safe region changes over time in the following two ways as shown in Fig. \ref{fig3}
\begin{equation}
    \begin{aligned}
        \mathbb{T}_{sc}:\mathbb{SR}^{i+1}(\boldsymbol{s}_{vk}) & =  \mathbb{SR}^{i}(\boldsymbol{s}_{vk}) - v_{max}t  \\
        \mathbb{T}_{se}:\mathbb{SR}^{i+1}(\boldsymbol{s}_{vk}) &= \min_{\boldsymbol{s}_{vj}\in \mathcal{N}(\boldsymbol{s}_{vk})}\mathbb{SR}^{i}(\boldsymbol{s}_{vj})  \\
    \end{aligned}\label{eq:safe_region}
\end{equation}
where $v_{max}$ represents maximum prior velocity in the environment and $\mathcal{N}(\boldsymbol{s}_{vk})$ represents the neighbourhood of $\boldsymbol{s}_{vk}$, i.e., $\boldsymbol{s}_{vj}$ satisfying the following equation
\begin{equation}
    | \boldsymbol{s}_{vk} - \boldsymbol{s}_{vj} | \cdot \mathbb{SR}^{i}(\boldsymbol{s}_{vj})<v_{max}t
\end{equation}
$\mathbb{T}_{sc}$ means that potential dynamic obstacles outside the safety region move radially in the direction of the vision state $\boldsymbol{s}_{vk}$ at maximum velocity. $\mathbb{T}_{se}$ represents the movement of obstacles outside the safety region perpendicular to the direction of the vision state. Thus, the evolution of the safe region is as follows:
\begin{equation}
    \mathbb{T}_{s}:\mathbb{SR}^{i+1}(\boldsymbol{s}_{vk})= \left\{\begin{matrix}
d_{max}  & v(\boldsymbol{s}_{vk}) = 1\\
\min{\left ( \mathbb{T}_{sc},  \mathbb{T}_{se}\right ) }  &v(\boldsymbol{s}_{vk}) = 0
\end{matrix}\right.
\end{equation}
where $v:\mathbb{S}^2 \mapsto \{\text{0}, \text{1}\}$ represents whether the direction of $\boldsymbol{s}_{vk}$ is visible or not.
\par Subsequently, we document the reason for each safe region evolution. This documentation allows the identification of the potential hazard location prior to the time evolution, leading to the generation of a potential obstacle, denoted as $\boldsymbol{s}_{eh}$. The position and velocity attributes of this potential obstacle are configured to pose a threat to the robotic arm at a specific point in time. In summary, the OUE law can be illustrated in the following form
\begin{equation}
    \mathcal{E}^{i+1}= f(\mathcal{E}^{i},\mathcal{O}) \cup\{\boldsymbol{s}_{eh}(\mathbb{SR}^{i},\mathcal{O})\}
\end{equation}
where $f(\mathcal{E}^{i},\mathcal{O})$ depicts the evolution of the known SCEs and $\boldsymbol{s}_{eh}(\mathbb{SR}^{i+1},\mathcal{O})$ represents the potential SCE generated by safe region.

\subsection{COOS search}
Based on the established obstacle modeling and evolution framework, we assess the probability of collision for the upcoming period to ascertain real-time collision risk. Suppose the ability to determine collision between SCE and robotic arm is established as $l: \boldsymbol{s}_{e}\times \boldsymbol{q}_{free} \mapsto \{\text{0, 1}\}$, where $l(\boldsymbol{s}^i_{ek}, \boldsymbol{q}^i) = \text{1}$ means \textit{k}-th SCE collides with robotic arms at \textit{i}-th time interval. In the following derivation, $l(\boldsymbol{s}^i_{ek}, \boldsymbol{q}^i)$ will be abbreviated to $l(\boldsymbol{v}^i_k)$. $l(\boldsymbol{v}^i_k)$ represents the velocity of the $k$-th SCE at $i$-th time. 
\par For each $\boldsymbol{s}_{ek}$ in $\mathcal{E}^i$, if the dynamic obstacle is assessed by the collision criterion $l(\boldsymbol{v}^{i+1}_k)$ to collide with the robot and the observation is an invalid observation, the CPE is assigned a value of 1. Conversely, if the observation is a valid observation, the CPE is reduced to $1-u$, which means the upper bound on the probability that the dynamic obstacle will not be observed is $1-u$. On the other hand, when $l(\boldsymbol{v}^{i+1}_k) = 0$, a more detailed evaluation ensues. If the obstacle accelerates at its maximum rate from the initiation of planning and collide with robotic arm, which means $l(\boldsymbol{v'}^{i+1}_k) = 1$, the CPE reverts to $1-u$. Conversely, if no collision occurs, the CPE is set to 0. Consequently, our collision probability of \textit{k}-th SCE is estimated as follows
\begin{equation}
    \begin{aligned}
        \hat{p}_{ck}^{i, i+1} &=(1-l(\boldsymbol{v}^{i+1}_k))l(\boldsymbol{v'}^{i+1}_k)(\mathcal{O}_{ek}-1)(u^{i+1}_{ek}-1)\\ 
        &+ l(\boldsymbol{v}^{i+1}_k)\left( -\mathcal{O}_{ek}u^{i+1}_{ek} +1  \right) \label{eq:PoE}
    \end{aligned}
\end{equation}
where $\boldsymbol{v'}^{i+1}_k = \boldsymbol{v}^{i+1}_k+\frac{a_{max}t^2}{2\|\boldsymbol{v}^{i+1}_k\|} \boldsymbol{v}^{i+1}_k$.

\par The preceding discussion addressed the computation of CPE within optimization problems. Subsequently, we will employ a Markov decision process to formulate and effectively resolve the optimization problem in (\ref{eq:optimal}).
\begin{figure}[!t]
    \includegraphics[width=9cm]{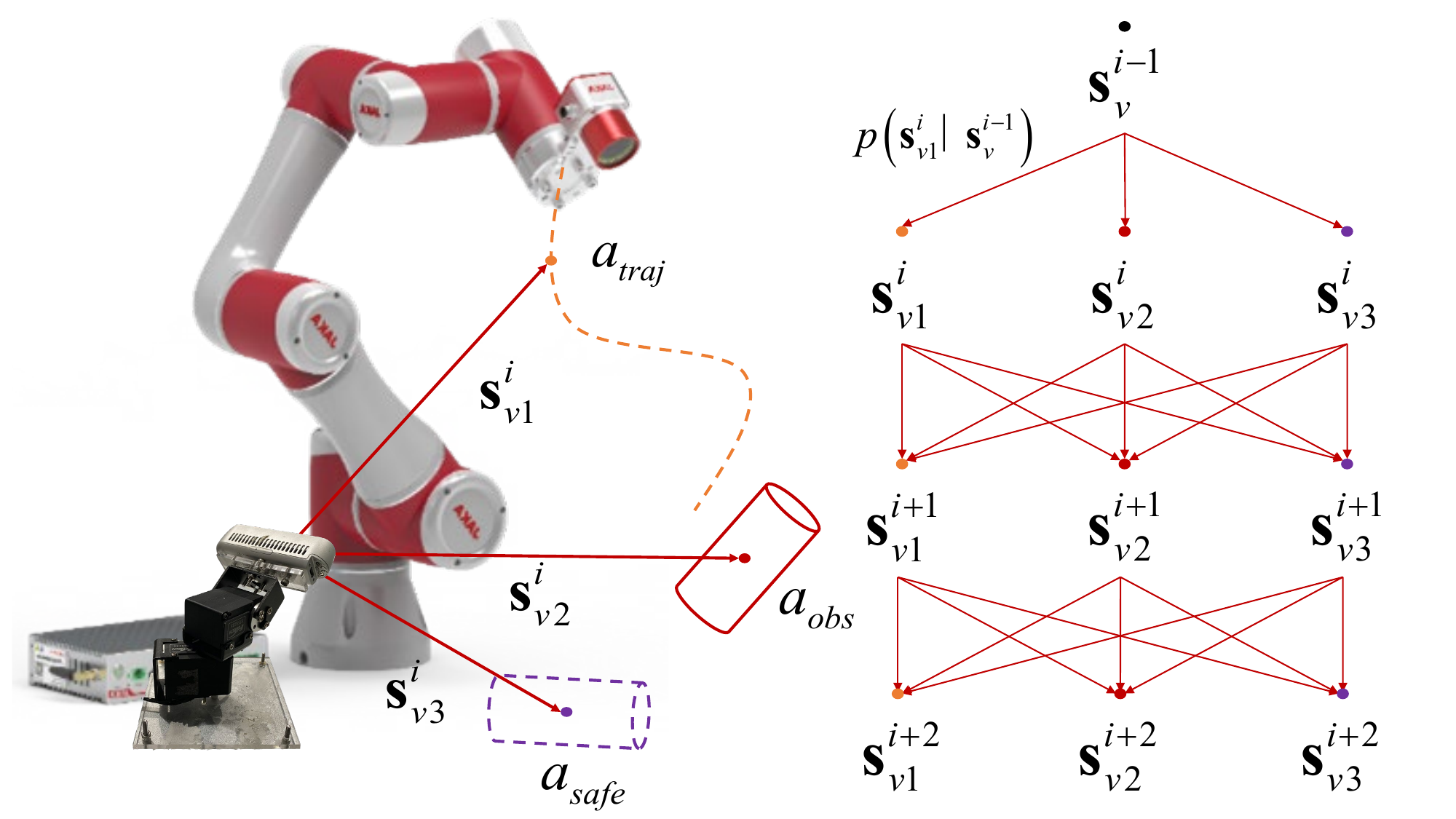}
    \centering
    \caption{Exposition of the COOS Search method: the left figure illustrates the generation of the action set derived from the robotic arm trajectory, considering both known obstacles and safe spaces; the right figure depicts Markov decision process search.}
    \label{fig2}
\end{figure}
\par A Markov decision process can be fully represented by a quaternion $\left \langle   \mathcal S,\mathcal A,P,R \right \rangle$. $\mathcal{S}$ represents the state set, $\mathcal{A}$ represents the action set, $P$ represents the probabilistic transition model, and $R$ represents the reward model. Let's begin by defining the action set, and the state set. Naturally, the states set can be defined equal to the vision state set. Given that the state of the active vision mechanism is known, We define the action set as a list of decisions, allowing us to opt for observing the end of the robotic arm, denoted as $a_{traj}$, observing previously estimated dynamic obstacles and potential obstacles generated from safe region, denoted as $a_{obs}$ and $a_{safe}$. After that, we need to determine the state transfer probability $p \left( \boldsymbol{s}^i_v \mid \boldsymbol{s}^{i-1}_v \right)$. To enhance the smoothness of the motion of the active vision mechanism, we define that the larger the difference between the front state and the back state on $\mathbb S^2$, the lower the probability that the active vision mechanism will transfer states. It's noteworthy that our optimization objective is also a probability. Therefore, it is advantageous to incorporate this state transfer probability directly into the optimization objective. The resulting optimization objective is given by
\begin{equation}
\begin{aligned}
\tau_{v,d} &= \arg _{\tau_v} \max \sum_{i=1}^n \gamma^i \ln \left(1-\hat{p}_{c}^{i, i+1}\right)\\
&+\alpha^i \ln \left(p\left(\boldsymbol{s}^i_v \mid \boldsymbol{s}^{i-1}_v\right)\right)
\end{aligned}
\end{equation}
which is the definition of $R$. To summarize, The following Alg. \ref{algorithm2} is a pseudocode illustration of COOS Serach.
\begin{algorithm}
    \caption{COOS Search}\label{algorithm2}
    \KwIn{$\boldsymbol{s}_{vi}$, $\mathbb{SR}^{i}$, $\mathcal{E}^{i}$,$\tau^i$, $t_p$, $n_p$}
    \KwOut{$\tau_v$}
    $\Delta t = \frac{t_p}{n_p}$\;
    $visionStateSample = genStates(\mathcal{E}^{i}, \mathbb{SR}^{i}, \tau^i)$\;
    $visionTrajList = genTrajs(visionStateSample)$\;
    $bestJ = 0$; $bestT = visionTrajList[0]$\;
    \For{each $\tau_{vk}$ in $visionTrajList$}{
        $\mathbb{SR}^{i}_{cl}, \mathcal{E}^{i}_{cl} \leftarrow clone(\mathbb{SR}^{i}, \mathcal{E}^{i})$\;
        $J=0$\; 
        \For{each $j$ in $n_p$}{
            $\mathbb{SR}^{i+j+1}_{cl} \leftarrow \mathbb{T}_{s}\left(\mathbb{SR}^{i+j}_{cl},\Delta t, \tau_{vk}(j\Delta t)\right)$\;
            $ \mathcal{E}^{i+j+1}_{cl} \leftarrow \mathbb{T}(\mathcal{E}^{i+j}_{cl}, \mathbb{SR}^{i+j+1}_{cl}, \mathcal{O} )$\;
            $\hat{p}_{c}^{i+j, i+j+1} \leftarrow calp(\mathcal{E}^{i+j+1}_{cl}, \mathbb{SR}^{i+j+1}_{cl}, \tau^i, \mathcal{O})$\;
            $J\leftarrow J+\gamma^j \ln \left(1-\hat{p}_{c}^{i+j, i+j+1}\right)+\alpha^j \ln \left(p\left(s^{i+j+1}_v \mid s^{i+j}_v\right)\right) $\;
        }
        \If{$J > bestJ$}{
            $bestJ \leftarrow J$\;
            $bestT \leftarrow \tau_{vk}$\;
        }
    }
    $\tau_v \leftarrow bestT$\;
\end{algorithm}
\par We begin by generating state sequences and trajectory sequences, as illustrated above in Fig. \ref{fig2}. Subsequently, we traverse all trajectories, conducting safe space predictions and SCE predictions based on the OUE law for each step in the trajectory. We also estimate collision probabilities based on the aforementioned predictions and (\ref{eq:PoE}). Finally, we calculate the cost, and the active vision mechanism selects the optimal trajectory for execution.

\section{EXPERIMENT}
\subsection{Experiment Setup}
In both simulation and real-world experiments, we employed the octree map \cite{octomap} for obstacle representation and utilized the FCL library \cite{FCL} to achieve rapid collision detection between the robot and the octree map. Subsequently, we adopted the Bi-RRT algorithm \cite{BIRRT} to generate obstacle avoidance paths for the robot. The entire system is integrated into the Robot Operating System (ROS), with the robot arm being simulated and visualized through Rviz and MoveIt!. 
\par We utilized the system depicted in Fig. \ref{fig4} for real-world testing. Our constructed collaborative robot experimental system comprises two active vision mechanisms and a JAKA$\textsuperscript{\textregistered}$ Zu7 robotic arm. The active vision mechanism is equipped with a Realsense$\textsuperscript{\textregistered}$ D435 camera with a 72° FOV. The two perpendicular servos in the active vision mechanism are FEETECH$\textsuperscript{\textregistered}$ servos equipped with encoders with a resolution of 0.088°. These servos are connected to the servo driver board via TTL protocol, which, in turn, is linked to the computer via a USB cable. Additionally, the computer is connected to the JAKA$\textsuperscript{\textregistered}$ robotic arm control cabinet using a network cable, enabling centralized control of both the active vision mechanism and the robotic arm. The computer's CPU is an i7-12900H, and it is equipped with an RTX-3070Ti graphics card.
\begin{table}
    \centering
    \fontsize{8}{10}\selectfont    
    \tabcolsep=2mm
    \caption{Temporal coverage simulation test results.}
\begin{tabular}{cc|ccccccc}
    \toprule
Num & Method & Body & RA & RH & LA & LH & Avg\\
\hline
\multirow{3}*{1}
& fixed  & 1.00 & 0.558 & 0.318 & 0.00 & 0.00 & 0.375 \\
~ & TCP  & 1.00& 0.742 & 0.673 & 0.907 & 0.652 & 0.795 \\
~ & \textbf{CEASE} & 1.00 & 1.00 & 0.549 & 0.931 & 1.00 & \textbf{0.896}\\
\multirow{3}*{2}
& fixed  & 0.617 & 0.451 & 0.228 & 0.108 & 0.078 & 0.296 \\
~ & TCP  & 0.913 & 0.614 & 0.435 & 0.766 & 0.587 & 0.663\\
~ & \textbf{CEASE} & 0.836 & 0.736 & 0.489 & 1.00 & 0.905 & \textbf{0.793} \\
\bottomrule
\end{tabular}\vspace{0cm}
\label{tab:Training_sizes}
\end{table}
\begin{figure}[!t]
    \includegraphics[width=9cm]{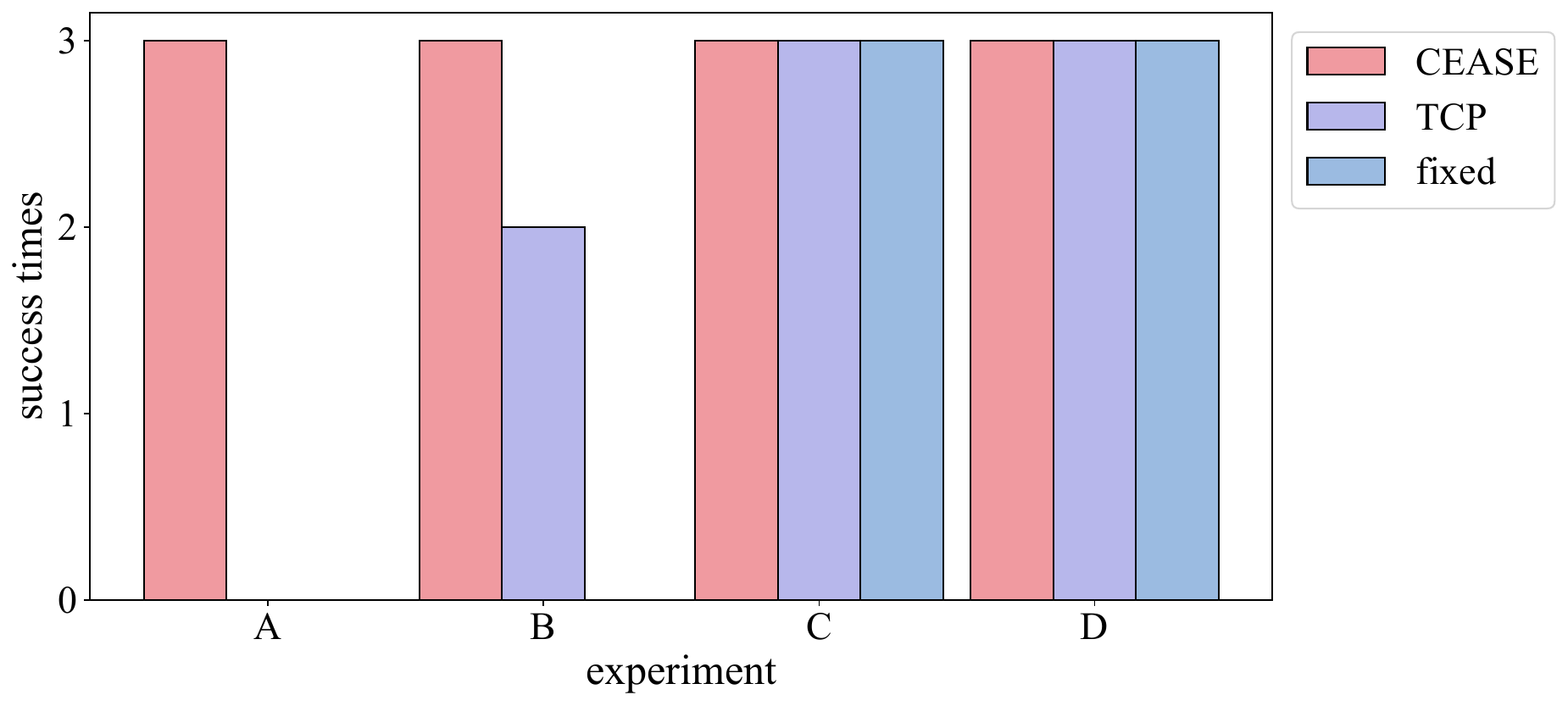}
    \centering
    \caption{Count of successful obstacle avoidance experiments for each method across various tests.}
    \label{fig:stra}
\end{figure}
\begin{figure}[!t]
  \centering
  \subfigure[]
  {\includegraphics[width=0.23\textwidth]{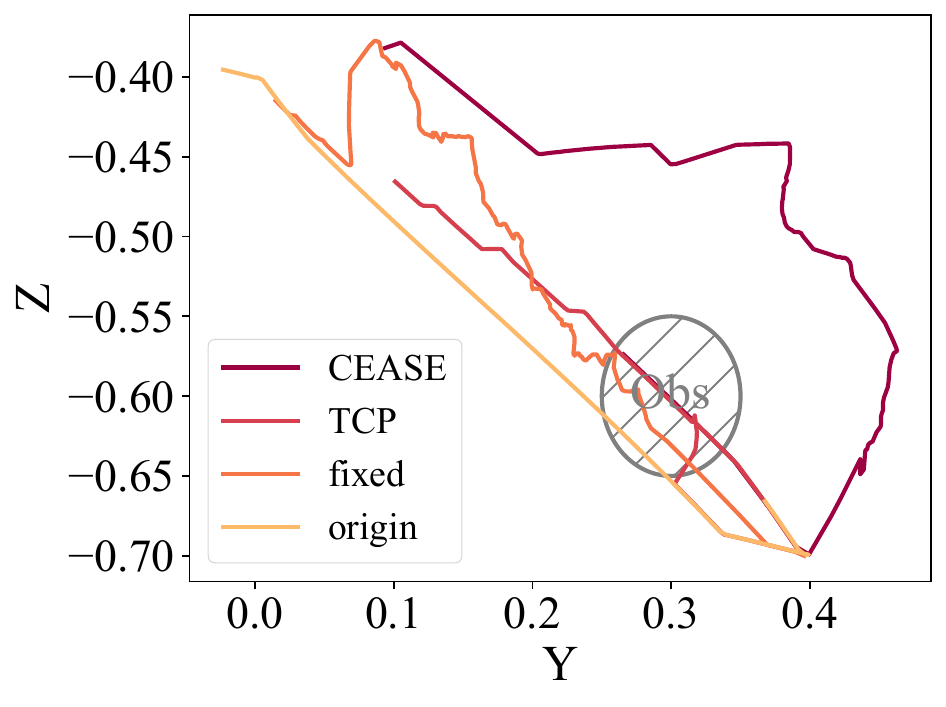}\label{fig:subfig7}}
  ~
  \subfigure[]
  {\includegraphics[width=0.23\textwidth]{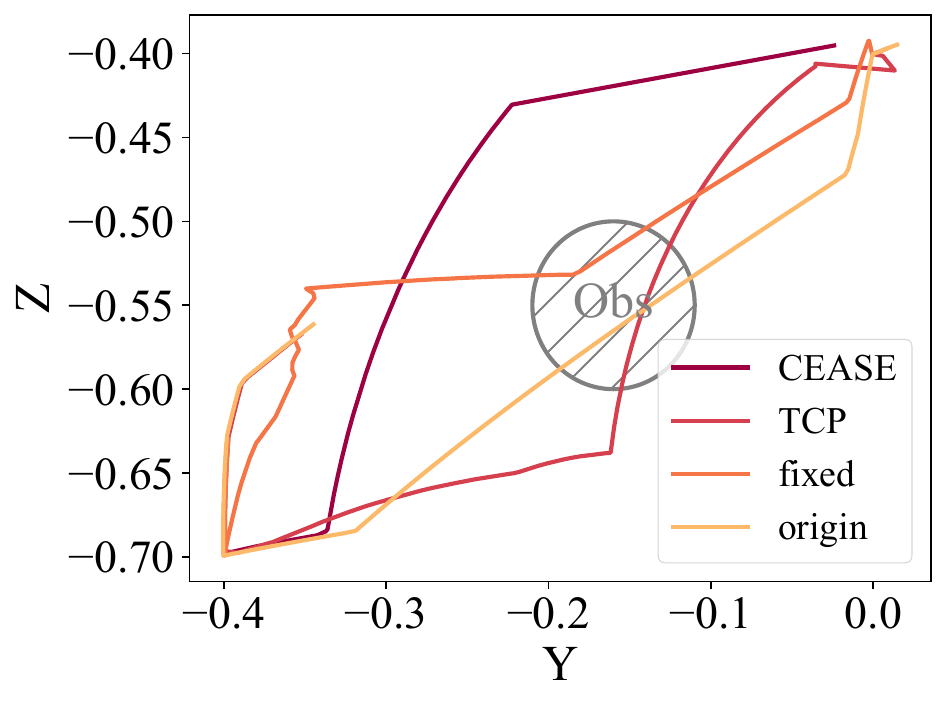}\label{fig:subfig8}}
   ~
  \subfigure[]
  {\includegraphics[width=0.23\textwidth]{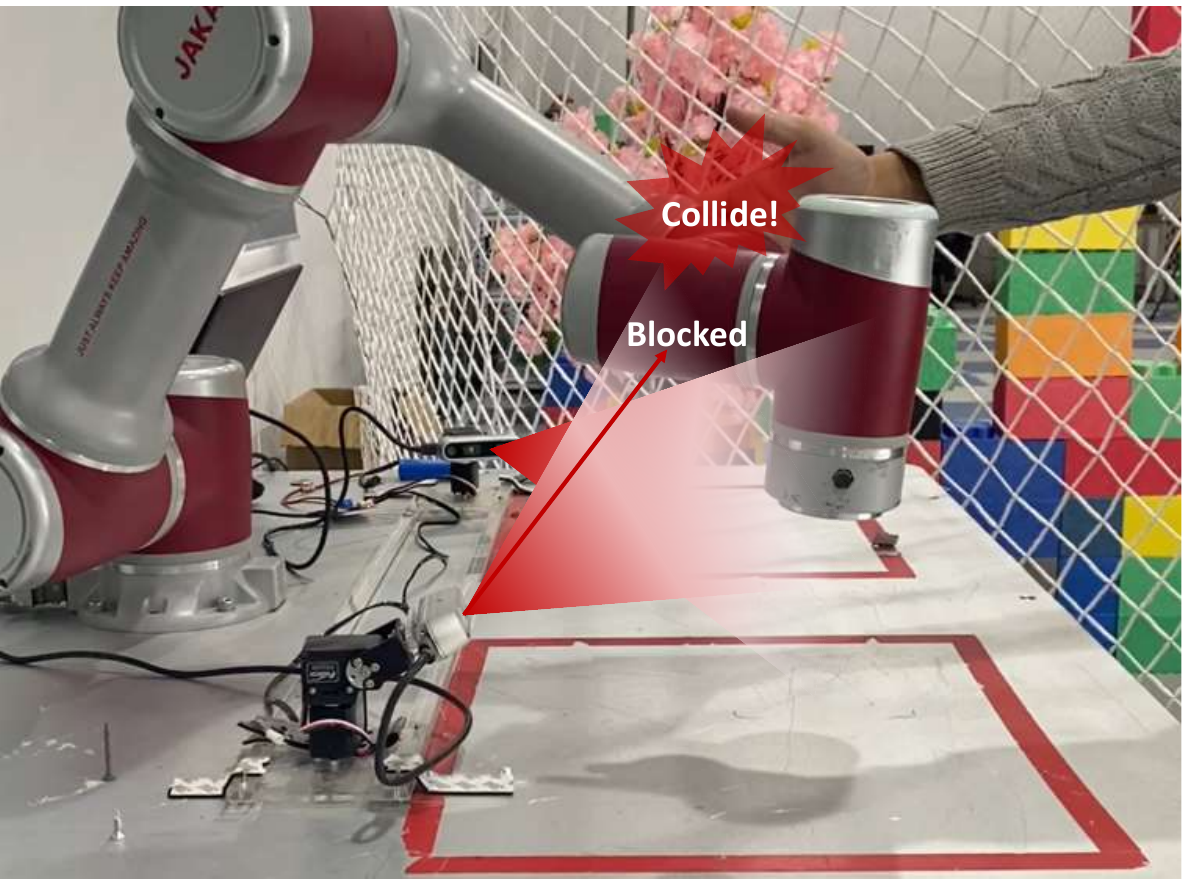}\label{fig:subfig9}}
~
  \subfigure[]
  {\includegraphics[width=0.23\textwidth]{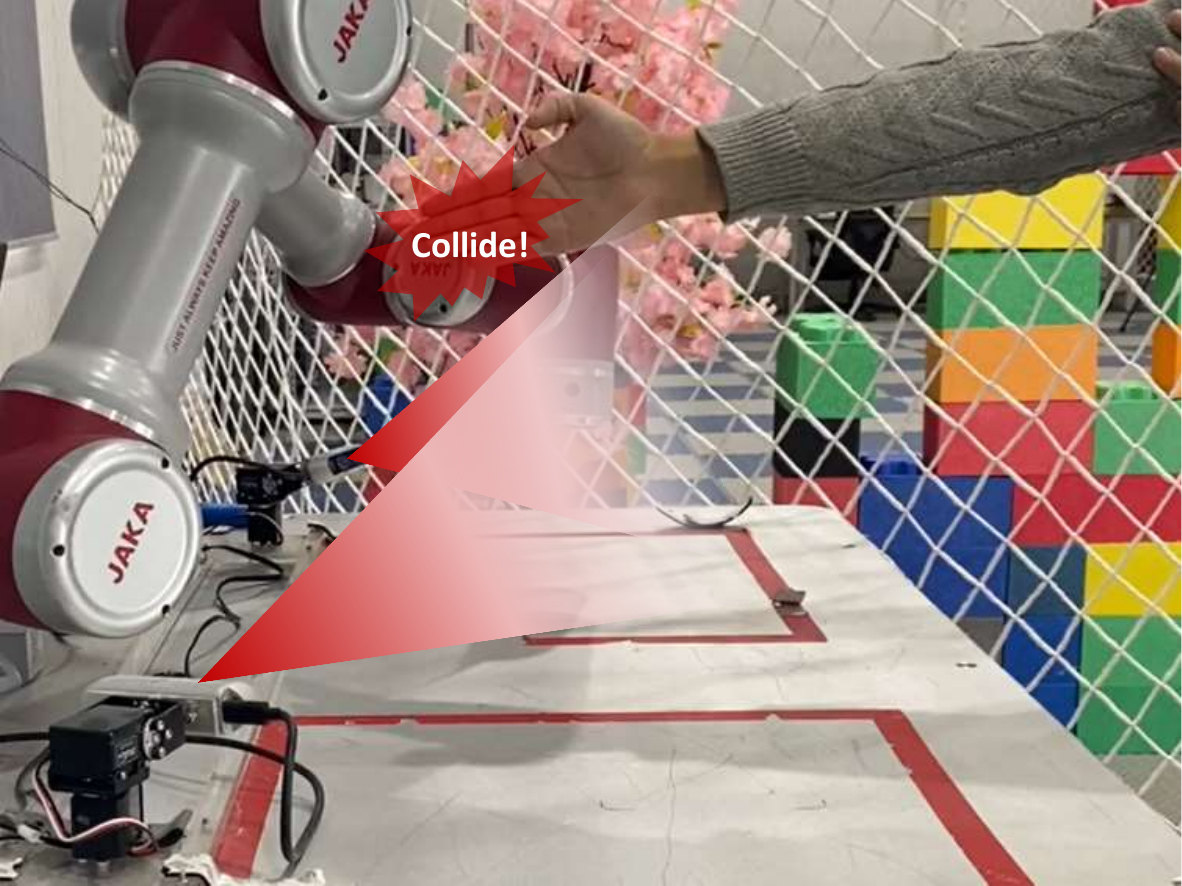}\label{fig:subfig10}}
~
  \caption{Static obstacle experiments: Figs. \ref{fig:subfig7} and \ref{fig:subfig8} depict individual methods alongside the original trajectories of the robotic arm in Experiment A and Experiment B. Figs. \ref{fig:subfig9} and \ref{fig:subfig10} illustrate the schematic representation of the TCP observation method, where the robotic arm collides with a static obstacle.}
\end{figure}\label{fig:real-world1}

\subsection{Simulation Tests}
Our experiments involves comparing our proposed method (CEASE) with an approach that solely focuses on tracking the robotic arm's endpoint (TCP) and another approach that fix the camera (fixed) at a specific vision state. The experimental conditions were standardized across all methods, including identical camera base placement, initial robotic arm trajectory, control algorithm, and dynamic obstacle trajectory.
\par We evaluated our method's performance through simulations in Rviz, set in two different working environments. In the first experiment, a humanoid obstacle was placed stationary in front of the robotic arm, with only its arm swinging at a predetermined rate. The second experiment modified this setup, allowing the simulator's body to move laterally. The robotic arm moves following a specified trajectory, and we assess the dynamic obstacle visibility by comparing the temporal coverage of our method with that of the TCP observation method and the fixed camera method for each joint of the human body. Temporal coverage is a ratio, with the numerator representing the duration during which a specific body part is concurrently observed by any camera within a period of time, and the denominator being the total time. The body parts include the following five elements: body, right arm (RA), right hand (RH), left arm (LA), and left hand (LH). The corresponding results are presented in Table \ref{tab:Training_sizes} and Avg means the average temporal coverage of the whole five parts.

\begin{figure}[!t]
    \includegraphics[width=9cm]{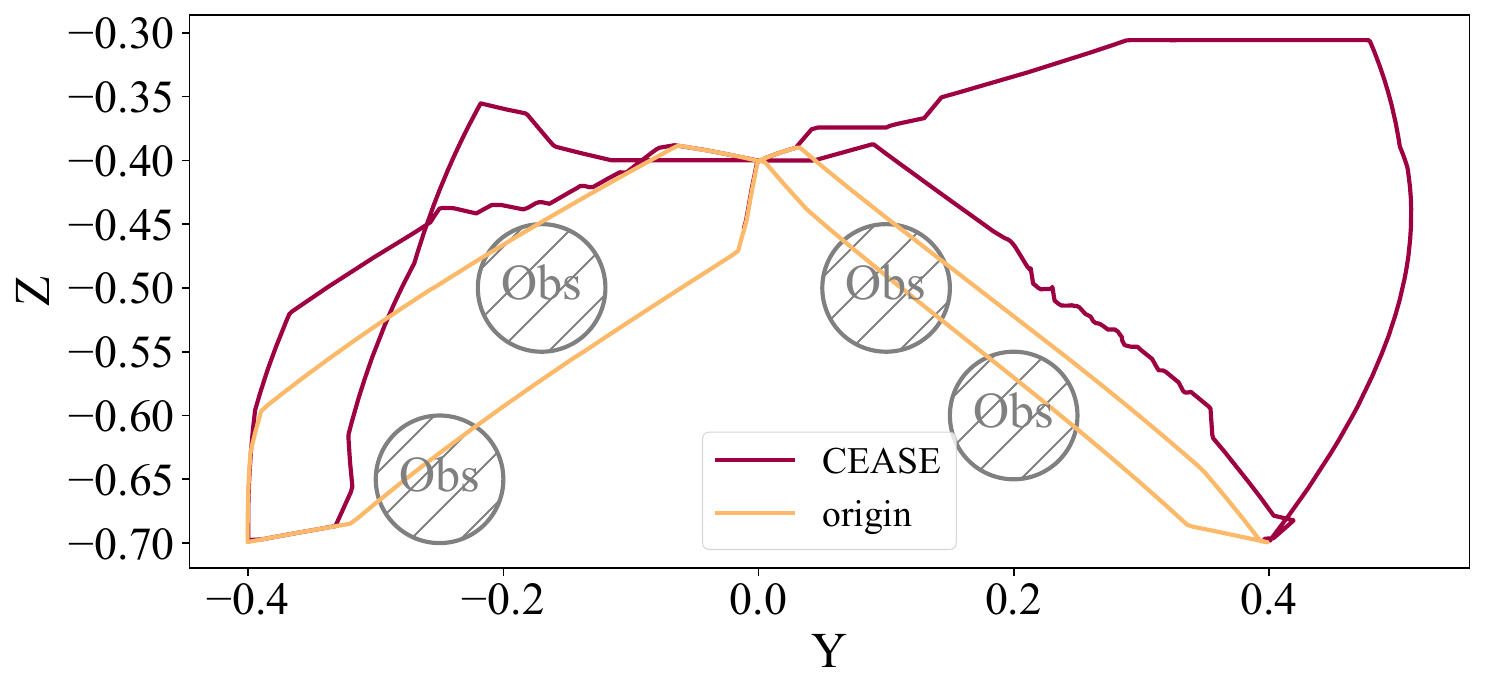}
    \centering
    \caption{The original trajectory and obstacle avoidance trajectory of the robotic arm using our observation method during operation.}
    \label{fig:Part1CEASE}
\end{figure}
\begin{figure}[!t]
    \includegraphics[width=9cm]{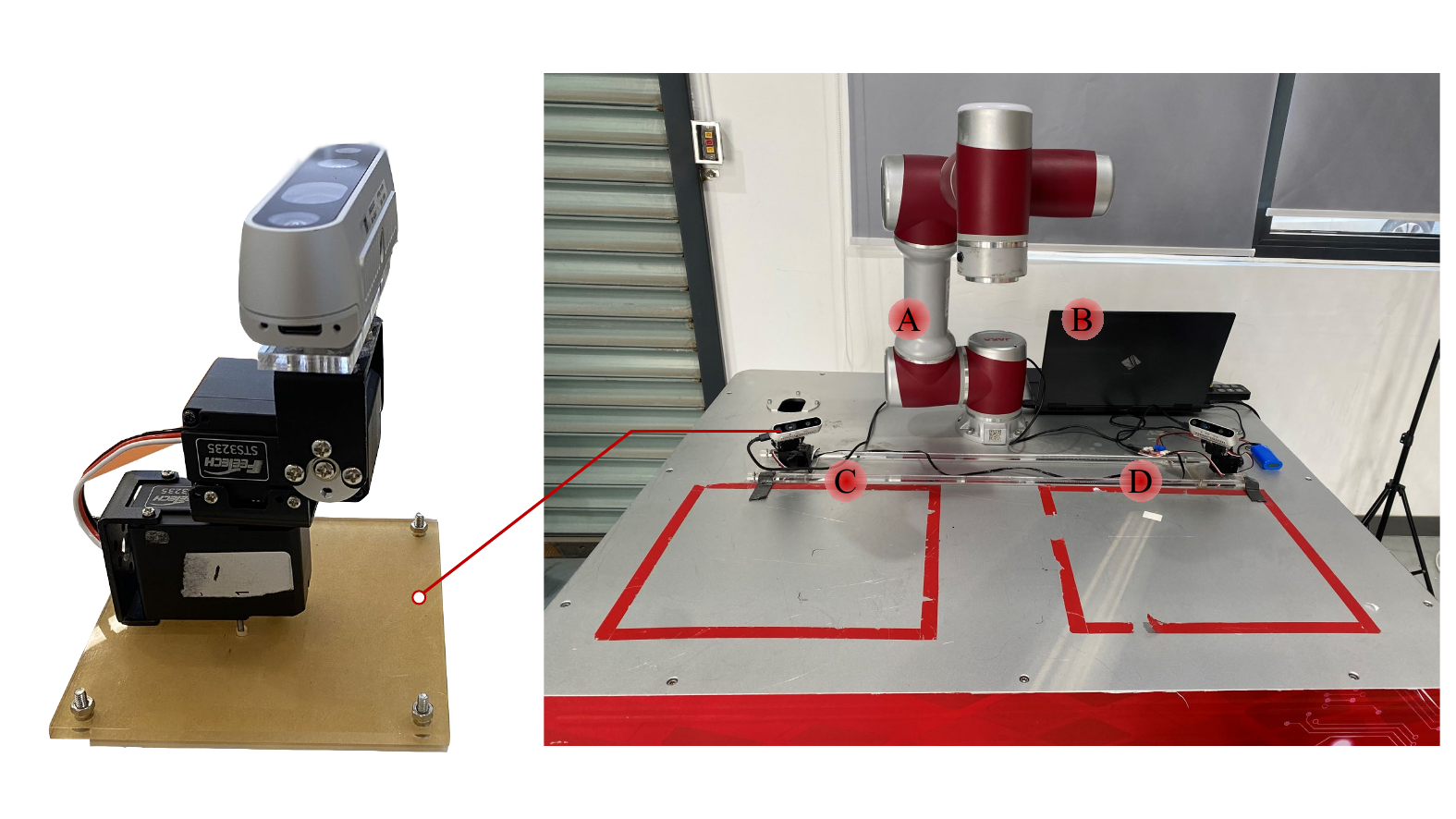}
    \centering
    \caption{Collaborative robotic arm equipped with an active vision mechanism. The active vision mechanism consists of two mutually perpendicular servos with RGB-D cameras.}
    \label{fig4}
\end{figure}
\par Table \ref{tab:Training_sizes} shows that, in terms of temporal coverage of human body parts, our algorithm outperforms both the fixed camera and TCP observation methods. In Experiment 1, the average temporal coverage of our method for each body part surpassed the fixed camera method by \textbf{139\%} and the TCP observation method by 12.7\%. In Experiment 2, these margins increased to \textbf{168\%} and 19.6\%, respectively. Our method's extension of observation beyond just the endpoint provides a more comprehensive observation of dynamic obstacles over time, enhancing the robotic arm's obstacle avoidance capabilities.

\subsection{Real-World Experiments}
\par We conducted two parts of experiments to validate the security and reliability of our system. The first part involved a static obstacle avoidance experiment. In this experiment, we deployed the active vision mechanism using three different methods: our proposed method (CEASE), observing the end of the robotic arm (TCP), and a fixed (fixed) method. We applied the same point cloud processing, octree construction algorithm, and robotic arm obstacle avoidance algorithm for all experiments. During the experiments, we positioned operator's hand at four different points shown in Fig. \ref{fig4} to assess the success rate of each obstacle avoidance method.We divided the experiments into four categories, labeled A to D, which were differentiated by the varying positions of the operator’s hand. For each category, three trials were conducted using each method. As illustrated in Fig. \ref{fig:stra}, our method consistently succeeded in avoiding the hand in all positions. Fig. \ref{fig:subfig7} and Fig. \ref{fig:subfig8} show the trajectories of the various methods in Experiment A and Experiment B versus the original trajectories. From Fig. \ref{fig:subfig9} and Fig. \ref{fig:subfig10}, It can be seen that the method of simply observing the TCP or fixing the camera angle can easily lead to obstacle avoidance failures due to the occlusion of the robot arm and the loss of the obstacle FOV. In contrast, Our approach precisely identifies the optimal observation direction, specifically the direction towards the operator's hand. Eventually, we positioned the operator's hands at four positions, A to D, within a robotic arm's operational cycle using our CEASE method for observation. The active vision system effectively detected all obstacles and executed obstalce avoidance trajectory. Fig. \ref{fig:Part1CEASE}  compares the robot arm's original trajectory (origin) with its obstacle-avoidance path (CEASE).
\begin{figure*}[!t]
  \centering
  \subfigure[$t=t_1$]
  {\includegraphics[width=0.17\textwidth]{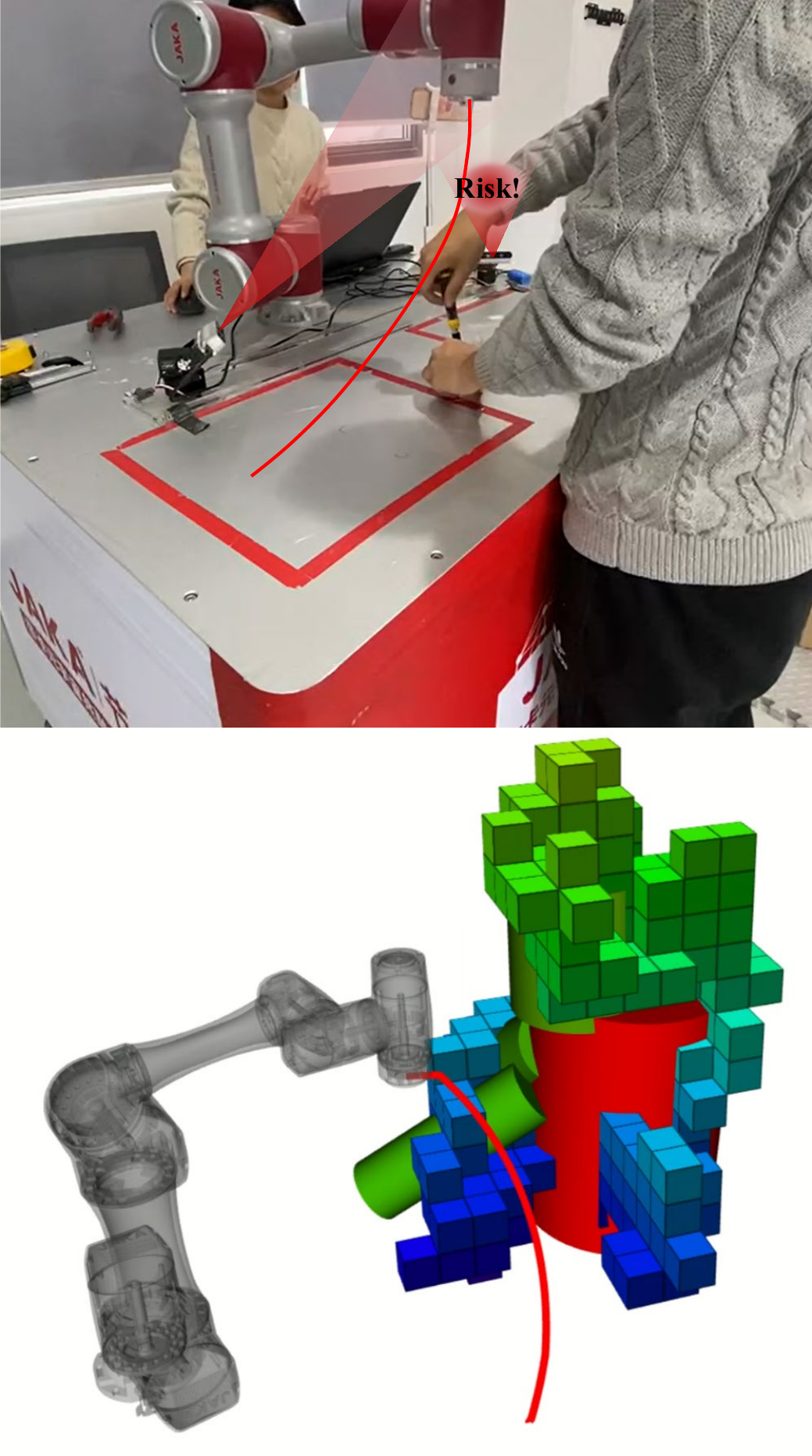}\label{fig:1r}}
  ~
 \subfigure[$t=t_2$]
  {\includegraphics[width=0.17\textwidth]{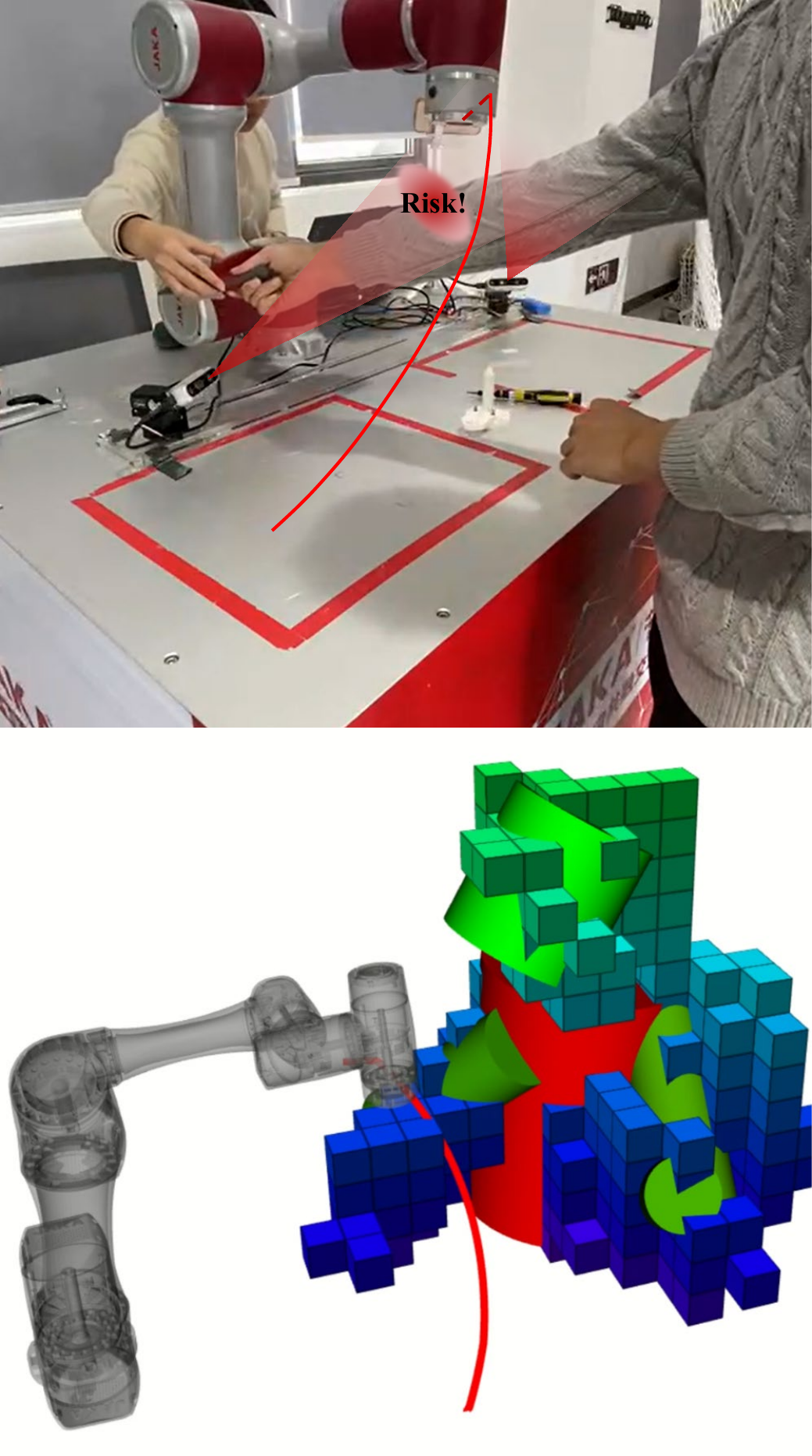}\label{fig:2r}}
   ~
  \subfigure[$t=t_3$]
  {\includegraphics[width=0.17\textwidth]{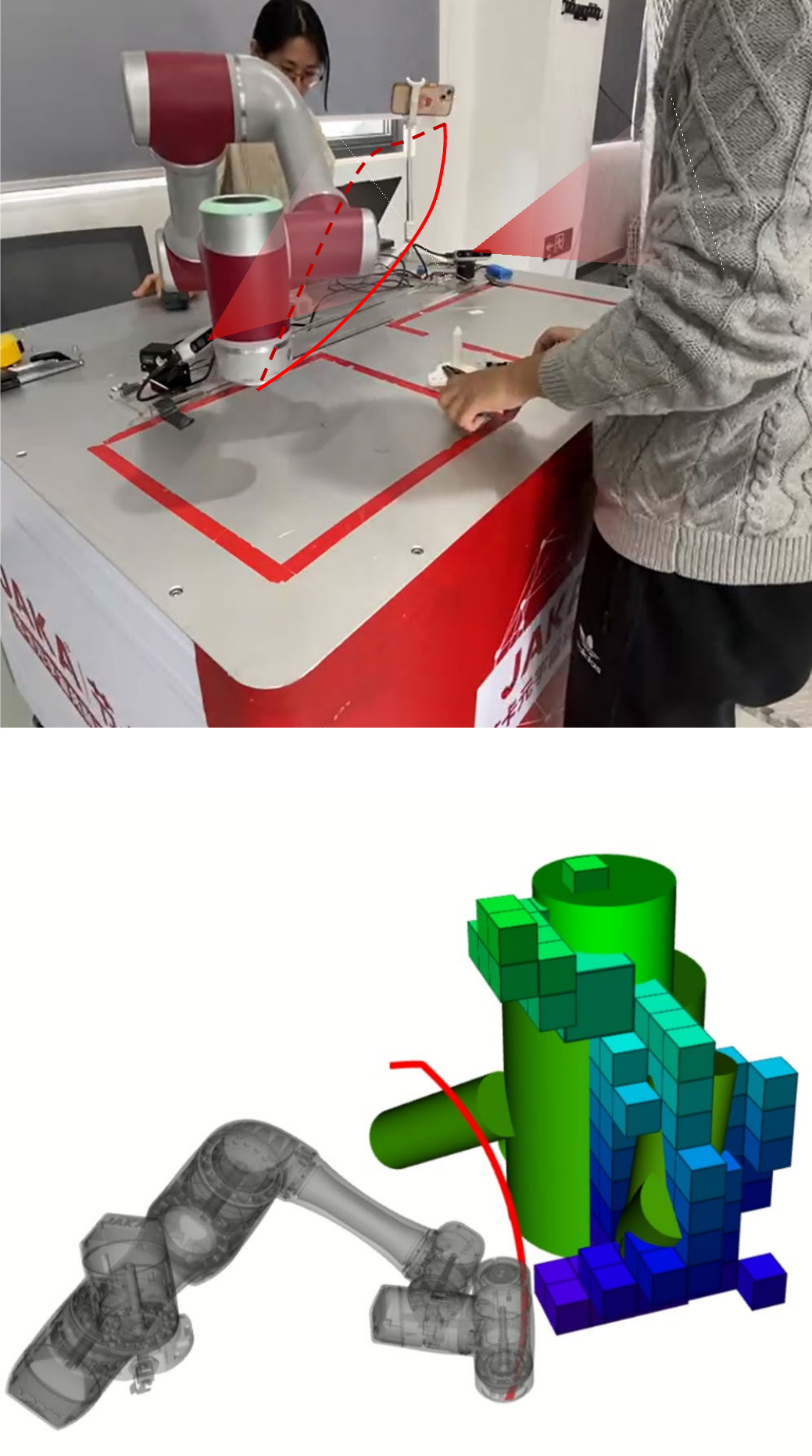}\label{fig:3r}}
   ~
  \subfigure[$t=t_4$]
  {\includegraphics[width=0.17\textwidth]{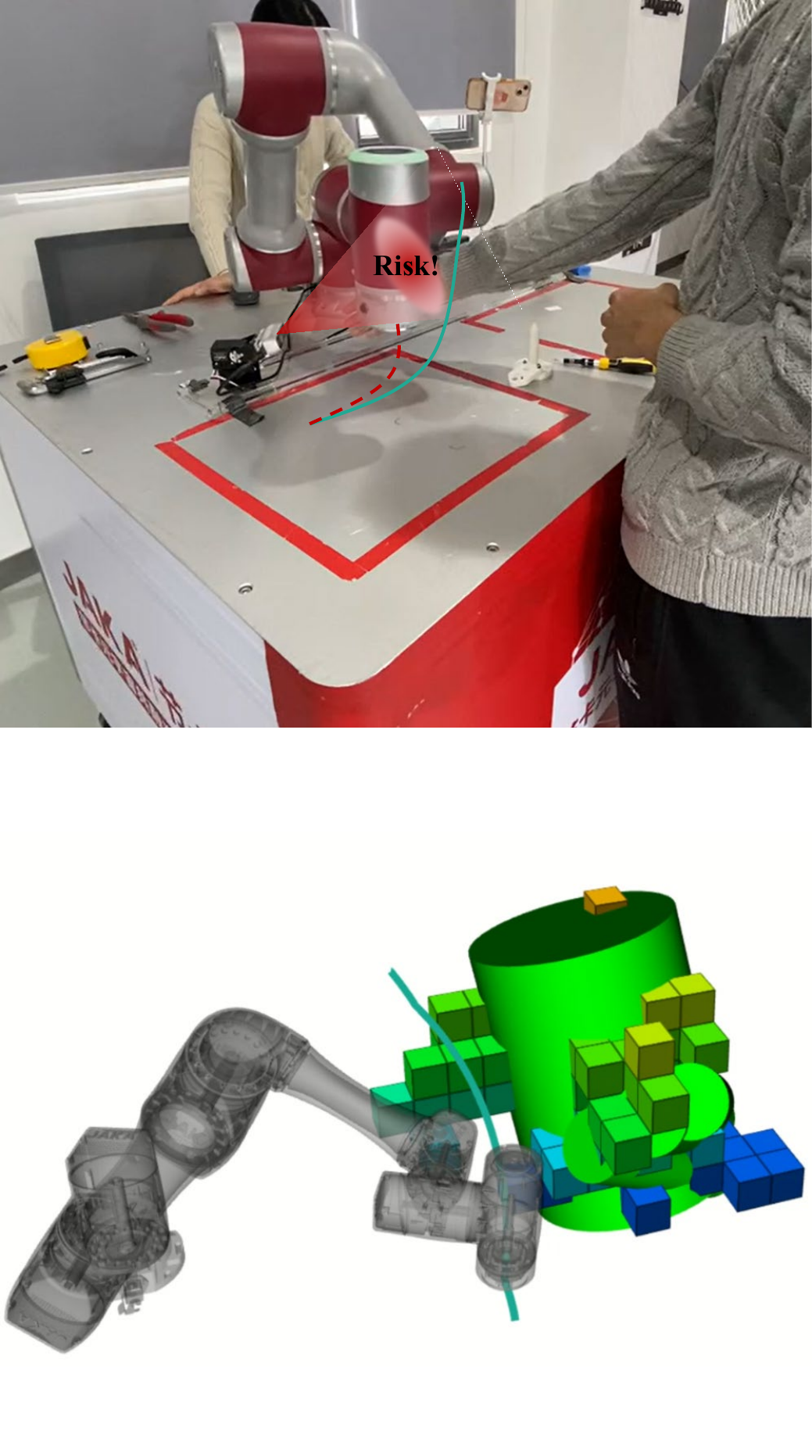}\label{fig:4r}}
   ~
  \subfigure[$t=t_5$]
  {\includegraphics[width=0.17\textwidth]{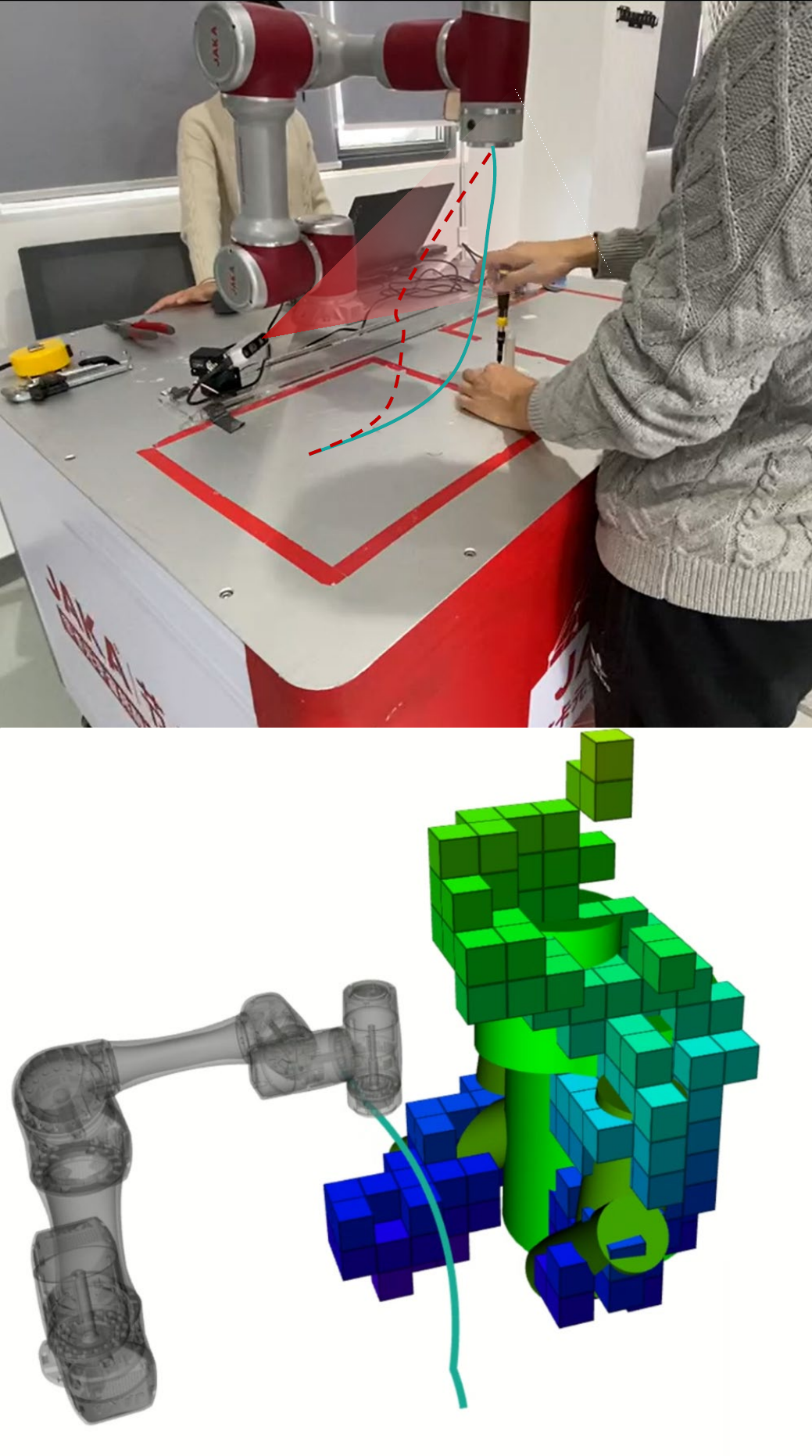}\label{fig:5r}}
    ~
  
  \caption{Obstacle avoidance experiments in real working scenarios: The upper figures in (a)-(e) illustrate the movement of a humanoid obstacle in a real environment. The solid red and green lines represent the original trajectory of the robotic arm, while the red dashed line indicates the trajectory during obstacle avoidance. The viewing angles of the two cameras are also annotated in the figures. The lower figures present data visualization in Rviz, where a more intense red in the SCE corresponds to a higher Collision Probability Estimate (CPE). The original trajectory is likewise depicted in the visualization.}\label{fig:real-world}
\end{figure*}
\begin{figure}[!t]
  \centering
  \subfigure[Robotic arm downward]
  {\includegraphics[width=0.235\textwidth]{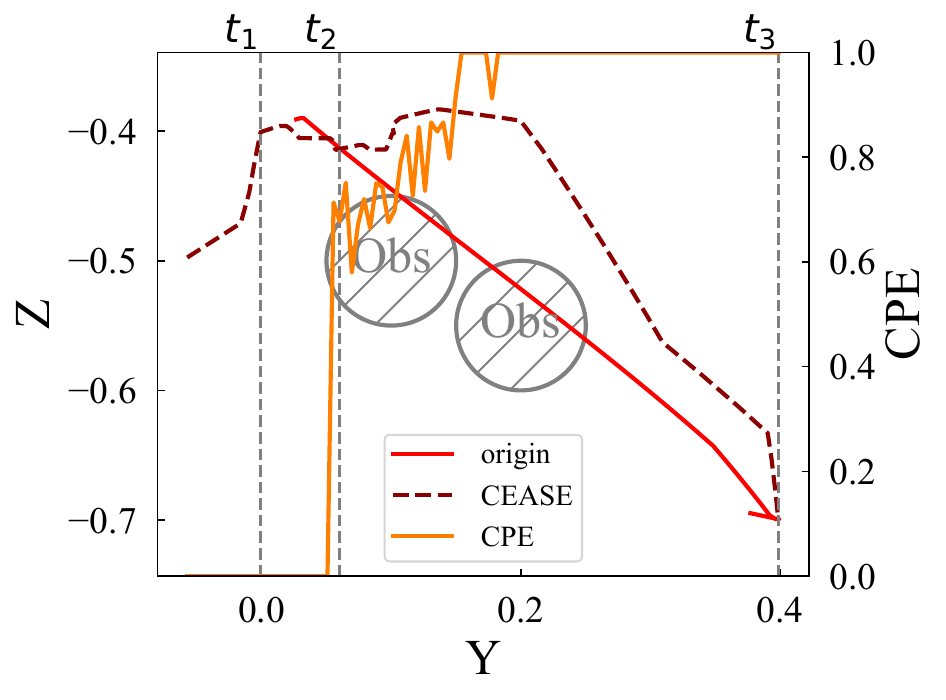}\label{fig:t1}}
~
  \subfigure[Robotic arm upward]
  {\includegraphics[width=0.235\textwidth]{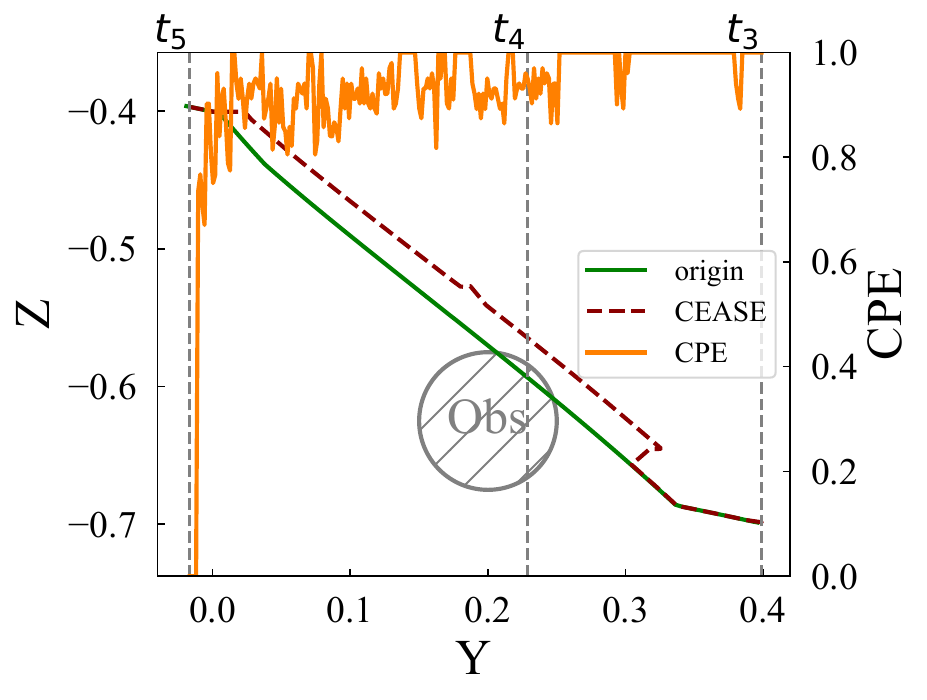}\label{fig:t2}}
  \caption{The trajectories of the robotic arm and CPE for the real working scenarios experiment. }\label{fig:real-worldt}
\end{figure}
\par In the other part of the experiment, we simulated the interaction between the operator and the robotic arm. As depicted in Fig. \ref{fig:real-world}, the operator was involved in machining a part within the robotic arm's workspace and engaged in the transfer of tools with another operator, potentially affecting the original trajectory of the robotic arm. During the experiment, when the robotic arm was ready to move downward, as illustrated in Fig. \ref{fig:1r}, the CEASE system detected that the operator's machining activity impacted the robotic arm's operation, prompting the robotic arm to perform its first replanning. Subsequently, as shown in Fig. \ref{fig:2r}, when the other operator handed a tool to the first operator, affecting the avoidance trajectory of the robotic arm, the CEASE system accurately identified movement of the obstacles. Then the robotic arm then executed the new obstacle avoidance trajectory until reaching the endpoint, as depicted in Fig. \ref{fig:3r}. Following this, when the operator finished using the previously used tool and decided to return it to the original position, this action once again interfered with the trajectory of the robotic arm, as shown in Fig. \ref{fig:4r}. Ultimately, the CEASE system successfully detected this action, and the robotic arm completed the execution of the obstacle avoidance path, as seen in Fig. \ref{fig:5r}. Visual representations of the obstacle at corresponding moments, CPE (where a redder color indicates a higher CPE), octree map, the original trajectory of the robotic arm, and the obstacle avoidance trajectory are provided below these figures. Additionally, these trajectories and CPE are presented more explicitly in Fig. \ref{fig:real-worldt}. 

\par According to simulations and real-world experiments, our CEASE system tracks dynamic obstacles swiftly, enhancing the safety of collaborative robotic arms. Our approach surpasses alternative methods in terms of observation range and temporal coverage of dynamic obstacles, thereby affirming our system's safety and robustness.

\section{CONCLUSION}
The challenge of collision detection in the interaction between a collaborative robotic arm and a human is of utmost importance. In addressing this issue, we present the CEASE system in this study. Distinguished from conventional depth camera setups, our CEASE system can be integrated onto the robot base without the need for additional arrangements, substantially reducing installation complexity. To validate the robustness of our system, we conducted comprehensive experiments, encompassing simulations and real-world scenarios. 
In simulation experiments, we confirmed that the CEASE system's efficacy in enhancing temporal awareness and ensuring a more comprehensive observation of dynamic obstacles. Compared to fixed cameras, our CEASE system demonstrated a significant improvement in the time coverage of dynamic humanoid obstacles, with an increase of 139\% and 168\%. Real-world experiments involved a comparison of various methods, affirming that the CEASE system accurately senses obstacles and attests to its reliability in human-robot interaction scenarios. In future work, we plan to use this system for further experiments on obstacle avoidance and coupled path planning, enhancing the safety of robotic systems.

\bibliographystyle{IEEEtran}
\balance
\bibliography{Documents}

\end{document}